\documentclass[letterpaper, 10 pt, conference]{ieeeconf}  

\usepackage{cite}

\usepackage{booktabs}
\usepackage{multirow}
\usepackage{amsmath}
\usepackage{amssymb}
\usepackage{pdfpages}
\usepackage{caption}

\usepackage{graphics} 
\usepackage{graphicx}
\title{\LARGE \bf
SIMMF: Semantics-aware Interactive Multiagent Motion Forecasting for Autonomous Vehicle Driving
}

\author{Vidyaa Krishnan Nivash and Ahmed H. Qureshi
\thanks{The authors are affliated with the Purdue University, West Lafayette, IN 47907, USA. Emails: \{krish255, ahqureshi\}@purdue.edu}
}

\IEEEoverridecommandlockouts
\begin{document}

\maketitle
\thispagestyle{empty}
\pagestyle{empty}

\begin{abstract}

Autonomous vehicles require motion forecasting of their surrounding multiagents (pedestrians and vehicles) to make optimal decisions for navigation. The existing methods focus on techniques to utilize the positions and velocities of these agents and fail to capture semantic information from the scene. Moreover, to mitigate the increase in computational complexity associated with the number of agents in the scene, some works leverage Euclidean distance to prune far-away agents. However, distance-based metric alone is insufficient to select relevant agents and accurately perform their predictions. To resolve these issues, we propose the Semantics-aware Interactive Multiagent Motion Forecasting (SIMMF) method to capture semantics along with spatial information and optimally select relevant agents for motion prediction. Specifically, we achieve this by implementing a semantic-aware selection of relevant agents from the scene and passing them through an attention mechanism to extract global encodings. These encodings along with agents' local information, are passed through an encoder to obtain time-dependent latent variables for a motion policy predicting the future trajectories.  Our results show that the proposed approach outperforms state-of-the-art baselines and provides more accurate and scene-consistent predictions. 

\end{abstract}

\section{INTRODUCTION}

Driving is a deceptively simple activity requiring continuous visual sampling, periodic control, and sustained attention \cite{CASTRO}. Humans simplify this task by sampling the environment for critical control cues, typically involving predictable extrapolations of well-learned circumstances. Human cognition can perform these perceptual estimations with high accuracy to ensure safe driving. According to \cite{Tyrrell}, two separate visual perceptual mechanisms exist to make good estimations. The first being the ambient mode, concentrates on factors such as spatial orientation and locomotion. It involves vehicle guidance using information gained from optical flow to derive estimations of vehicle positioning. The second being the focal mode, focuses on object recognition, identification, and the semantics associated with it. It concentrates on the detection and identification of objects of importance in the environment such as physical barriers, and roadway threats, and uses its practical semantic associations as critical cues for deriving estimations. 

The current work in designing autonomous driving motion prediction methods focuses explicitly on the ambient mode and takes only implicit cues from the second, i.e., the focal mode, by inherently learning from data. Hence, the lack of context from the explicit modeling of the focal mode and its inherent interaction with the ambient mode limits the existing methods' performance accuracy and efficiency in complex environments. Therefore, in this paper, we present SIMMF, an extensible approach to introduce the focal mode and imbibe an active interaction between the two modes of perceptual mechanisms. Our key contributions are three-fold: Firstly, we formalize focal mode by formulating an algorithm to extract the semantics of agents and scene elements from environment maps. This extracted information is then utilized to optimize the grouping of relevant agents in a scene for their scene-consistent motion forecasting. Secondly, we combine both modes of perceptual systems through an attention mechanism to obtain their global, interactive representation. Thirdly,  we capture inter-group agents' interactions and new relevances in the data in a time-series manner using temporal encodings of focal and ambient representations. Finally, the motion prediction policy uses these temporal encodings to predict agents' future trajectories under scene consistency. Hence, SIMMF aims to inculcate an active interaction between the two visual systems and induce an ability to complement one another, achieving mutual reinforcement for trajectory predictions. This results in an overall reduction in response time and improves the accuracy of trajectory predictions compared to state-of-the-art (SOTA) baseline methods. 

\section{Related work}

Research in trajectory prediction has explored multiple facets of driving, uncovering a broad spectrum of approaches and possibilities.  Early works majorly delved into decision-making processes based on deterministic models like Dynamic Bayesian Networks \cite{Lefvre2014ASO}, Gaussian
 Process Regression (GPR) \cite{4359316} and Recurrent Neural Networks (RNNs) \cite{article} \cite{8460504} \cite{jia2022towards}. According to the review papers \cite{Leon_2021} \cite{Rudenko_2020}, motion prediction techniques can be classified based on the number of modes, level or type of abstraction, and method of approach. 
Many data-based approaches in the literature rely on LSTMs
or simpler CNNs
\cite{nikhil2018convolutional} to learn from agent states and histories. But these techniques do not ensure scene consistency, i.e., the trajectory predictions of multiple agents collide with each other or with static obstacles. 
To resolve this issue, recent studies have employed conditional variational autoencoders (CVAEs) or more recent methods
such as generative adversarial networks (GANs) and attention mechanisms to achieve SOTA work in motion prediction for autonomous vehicles. 

However, these techniques have time-independent latent variables and hence do not capture temporal dependencies within each agent's history and dynamic relationships between agents. For example, in CVAE, the latent variable encodings for the input agent states and histories are constant with respect to time. Hence they do not explicitly learn the temporal graph governing the causal relationships in its inputs. The papers \cite{7796911} \cite{girgis2022latent} state this problem for general time-series data and proposes temporally dynamic latent variables to account for the lag in its memory states.  

Some authors tend to resolve this issue using RNN for encoding the temporal information \cite{jain2016structuralrnn} \cite{jia2022towards}, but these methods cannot handle multimodal trajectory predictions. Similarly, \cite{lanegraph} predicts trajectories based on lane graphs by exploring both lateral (e.g.: lane keeping, turning) and longitudinal (e.g.: acceleration, braking) uncertainties. However, they do not capture the interaction between the two modes. Also, such works \cite{laformer} focus on lane semantics, thus the other semantics (roadblocks, pedestrian walkways, barriers) are left unexplored.

Another recent and perhaps the most relevant method to our approach is ScePT \cite{chen2022scept} framework, which follows an ambient mode of perception, ensuring scene consistency and computational efficiency. It is a discrete CVAE model that outputs joint trajectory predictions and achieves scene consistency using an explicit collision penalty as regularization during training. Furthermore, it reduces the computational complexity by forming subgraphs of agents, also known as cliques, based on Euclidean distance and jointly predicts their future states rather than for an individual agent or a whole set of agents. Although ScePT provides SOTA performance, it lacks the following features that limit its performance. First, it does not account for the temporal dependencies of the agents, which, as highlighted earlier, is crucial for better performance. Second, it only considers the ambient mode of perception and does not have a focal mode based on agents' local perception fields. Lastly, it forms cliques based on Euclidean distance only and does not account for other scene semantics such as lanes, road barriers, etc. Therefore, in this paper, we propose SIMMF, a full-stack motion prediction model that, with ambient and focal modes of perception, considers all possible scene semantics and temporal dependencies of selected agents, as summarized in Table~\ref{checkmark_tab}, and provides improved performance among all baselines.
\begin{table}[htbp]
\begin{center}
\begin{tabular}{cccccc}
\toprule Method & $\mathrm{SC}$ & $\mathrm{TE}$ & $\mathrm{AF}$ & $\mathrm{SI}$ & $\mathrm{OC}$  \\
\midrule $\begin{array}{l}\text {SRNN}[13]
\\
\text {SSTran}[12] \\
\text {LaFormer}[15]\\
\text {LaneGraph}[14] \\
\text {ScePT}[16] \\
\text{SIMMF}\end{array}$ & $\begin{array}{l}\\
\checkmark \\
\\
\\
\checkmark \\
\checkmark
\end{array}$ & 
$\begin{array}{l}
     \\ 
    \checkmark \\
    \checkmark \\
    \checkmark \\
    \\
    \checkmark
     
\end{array}$
& $\begin{array}{l}
\\
 \\
 \\
\\
\\
\checkmark
\end{array} $
& $\begin{array}{cc}
   \\
   \\
   \checkmark\\
  \checkmark \\
   \\
   \checkmark
\end{array}$
& $\begin{array}{l}
\checkmark \\ \\
\\
\\
\checkmark
\\
\checkmark\end{array}$\\
\bottomrule

\end{tabular}
\end{center}\vspace{-0.15in}
\caption{Legend: SC - Scene Consistency, TE - Temporal Encoding, AF - Ambient and focal mode interaction, SI - Semantic-aware Interaction, OC - Optimal computation time}\label{checkmark_tab}
\end{table}\vspace{-20pt}
\section{Proposed Method}
Our work presents a method to generate joint trajectory predictions of all the interacting agents in a scene. These agents are captured as a spatiotemporal graph $G^t = (V,E)$ for each timestep $t$ from the scene, where $V$ represents the nodes and $E$ represents the graph's edges. The agents are represented as nodes, and the interactions between them are represented by the edges. 
Henceforth, the terms nodes and agents are used interchangeably and are broadly classified into two types, namely vehicles and pedestrians utilized throughout the model to determine the dynamics $\psi$ of each agent based on its respective semantics. The state of a node is represented by $s\ \epsilon\ \mathbb{R}^4$ containing information about the position $p = (x,y)$, orientation $\theta$, and velocity ($v_x$, $v_y$).
Specifically, $s_v$ contains vehicle state information $s_v = (x,y, \nu, \theta)$,  where $\nu$ is the $L_2$ norm of $(v_x, v_y)$ and $s_p$ contains pedestrian information $s_p= (x,y,v_x,v_y)$. The states are stacked together from past $H$ timesteps to timestep $t$, given by $\vec h^t_s = (s^{t-H}, s^{t+1-H}, ..., s^t)$ while $\vec h^t_e$ represents the corresponding stack of edge information. We use the state information to model an edge between a pair of nodes. The edge quantifies the degree of interaction ($\alpha_{ij} \in [0,1]$), where $(i,j)$ denotes the pair of nodes.  

Our work assumes that maps containing geometric and semantic information are available. These maps are converted into local maps around each node. For a particular pixel resolution, we consider a corresponding fixed $K \times K$ size for all local maps, each modified into having $L$ semantic channels given by $M^t \ \epsilon \ \mathbb{R}^{K \times K \times L} $. The map semantic channels are constituted by the token identifiers $\tau$. For example, roadblocks are defined by a set of edge line tokens $\tau_{e_{0:N}}$ where $N$ represents the total number of tokens required to define the spline. Similarly, lanes are represented by curve spline tokens $\tau_l$. We form cliques $C^t$, i.e., subgraphs of agents having a high $\alpha_{ij}$ for a set of sampled timesteps $t \  \epsilon \  (0,T)$, $T$ representing the total number of timestamps considered for clique formation. We form the cliques based on the state and edge histories 
 along with map semantics given by $ C^t_{1:n} = \sigma(s_{t_{1:N}},\vec h^t_s, \vec h^t_e, M^t )$, where $n$ represents the number of cliques formed from $N$ number of agents in the scene. The function $\sigma$ leverages various road semantics from $M^t$, at time $t$, such as lane barriers, crosswalks, lane directions, the distance between agents, etc., to quantify the edge value $\alpha$ between nodes and forms cliques of nodes with $\alpha$ greater than a threshold. Finally, given a clique, $C^{t:t-H}$, agents' states, $h^t_s$, and interactions, $h^t_e$, at time $t$, we aim for a trajectory prediction function that predicts the future trajectories of each agent $s_{pred}^{t:t+\eta}$ for a horizon length $\eta$ based on combined measures of the focal and ambient mode of visual perception.
\begin{figure*}
\centering
\includegraphics[width=1.0\linewidth]
   {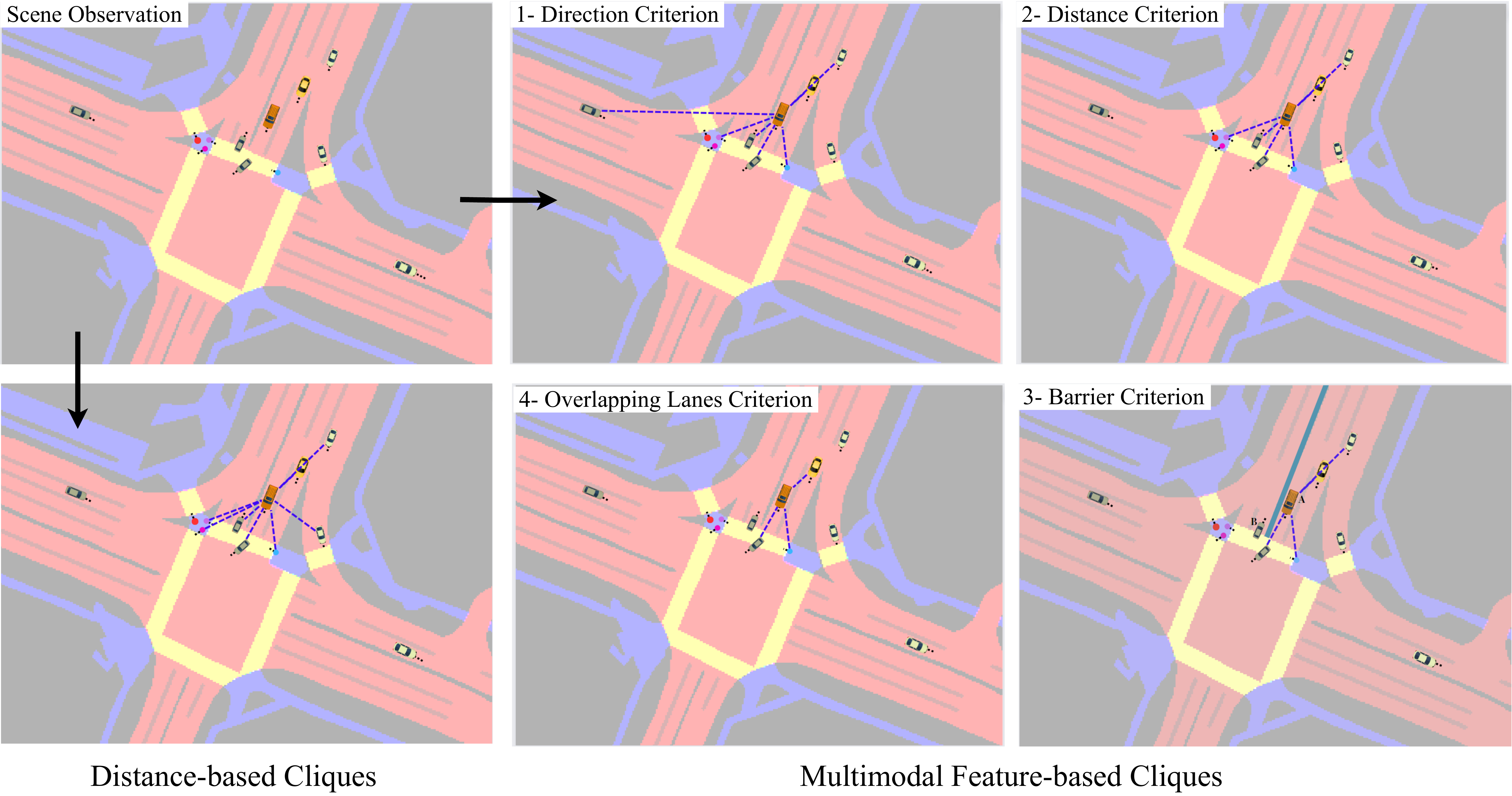}
   \vspace{-0.2in}\caption{ Shows a step-by-step illustration of how the cliques are pruned using semantic information in order to capture the focal mode of visual perception. The orange vehicle represents the ego-vehicle ($A$), differently colored circles represent pedestrians, and the cars represent other vehicle agents. Fig ~\ref{fig:clique}(1) shows the direction criterion effectively pruning agents which move away from the ego-vehicle. Fig ~\ref{fig:clique}(2) shows how cliques are formed based on a distance threshold from $A$.  Fig  ~\ref{fig:clique}(3) represents the barrier criterion pruning agent $B$ based on road semantics. Fig ~\ref{fig:clique}(4) illustrates how the lane semantics affect the clique formation.}
    \label{fig:clique}
\vspace{-0.2in}\end{figure*}
\subsection{Semantics-aware Clique Formation}
The task of predicting the trajectories of all agents in a scene  without collisions is complicated due to factors such as varying entry and exit timestamps and the interdependence between the multi-modal trajectories of different agents. To ensure scene consistency, each agent should be a part of the joint latent distribution, than individual predictions independent of its neighbors i.e. each agent should be associated with a discrete latent variable with cardinality $N$. Consequently, the cardinality $N$ grows exponentially with the number of nodes in the scene. To solve this increase in complexity, the scene can be broken down into smaller groups, called cliques formed based on degree of interaction. Fig \ref{fig:clique} shows techniques to form cliques based on various criteria, such as the distance, direction, and semantics between the agents. These techniques prune the cliques to include only the most relevant agents for decision-making, thereby eliminating cliques that contain redundant information.
\subsubsection{Euclidean distance parameter} 
Given a scene with multiple nodes, a spatio-temporal scene graph is generated where nodes represent agents and edges represent the interactions between these agents \cite{chen2022scept}. Each agent’s closest future distance is ascertained as a measure of interaction, propagating forward each node $i$, according to an action $a_i$ with a constant velocity model $\phi_{a_i}^t$. The $\phi_{a_i}^t$ maps the initial state of agent, $i$, to $t$ time steps ahead into the future states based on $a_i$ executed at the initial state, at time $t=0$. 
The closest future distance between two agents is defined as $d_{ij} = \min ||\phi_{a_i}^t(s_i), \phi_{a_j}^t(s_j)||_2$. We then define $\alpha_{ij}$ to populate the scene graph adjacency matrix as follows: the $\alpha_{ij} =0$ when $d_{ij} > d_0$ and $\alpha_{ij} =\frac{d_0}{d_{ij}}$ when $d_{ij} < d_0$, where $d_0$ is the threshold distance fixed according to the agent type. 
     
\subsubsection{Interaction detection using direction}

To ensure that the cliques do not include agents with minimal or no interaction, we categorize the agents based on whether they are moving toward or away from each other.
For example, even if agents are within a distance threshold $d_0$, they will have no impact on each other if they are moving in opposite or perpendicular directions. One way to compute this is by taking the ego agent's viewpoint as the observer having position and velocity $(p_o, v_o)$ and determining $(p_i,v_i)$ of the other agents relative to the observer's frame of reference. 
The interaction between the ego vehicle and any other agent is considered relevant when there is a timestep within $\eta$ such that the two velocity vectors intersect or lie within a relative distance threshold $D$ \cite{herencia2010formal}. Formally, 
$(p_o + t v_o) - (p_i + t v_i) = \vec 0 {\iff} V_o \cap V_i$. Since, $ (p_o + t v_o) - (p_i + t v_i) = (p_o - p_i) + t (v_o - v_i),$ 
we check if the $||(p_o - p_i) + t (v_o - v_i)||_2 < D$. This technique also implicitly ensures scene consistency by introducing direction-based collision checking during clique formation. 
\begin{figure}
    \centering
    \includegraphics[width = 0.23\textwidth, height=5.21cm]{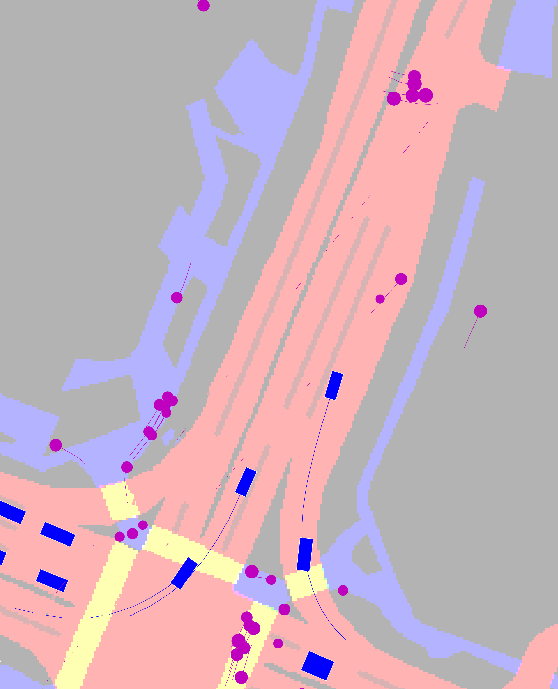}
     \includegraphics[width = 0.23\textwidth, trim={0 0 0cm 0},clip]{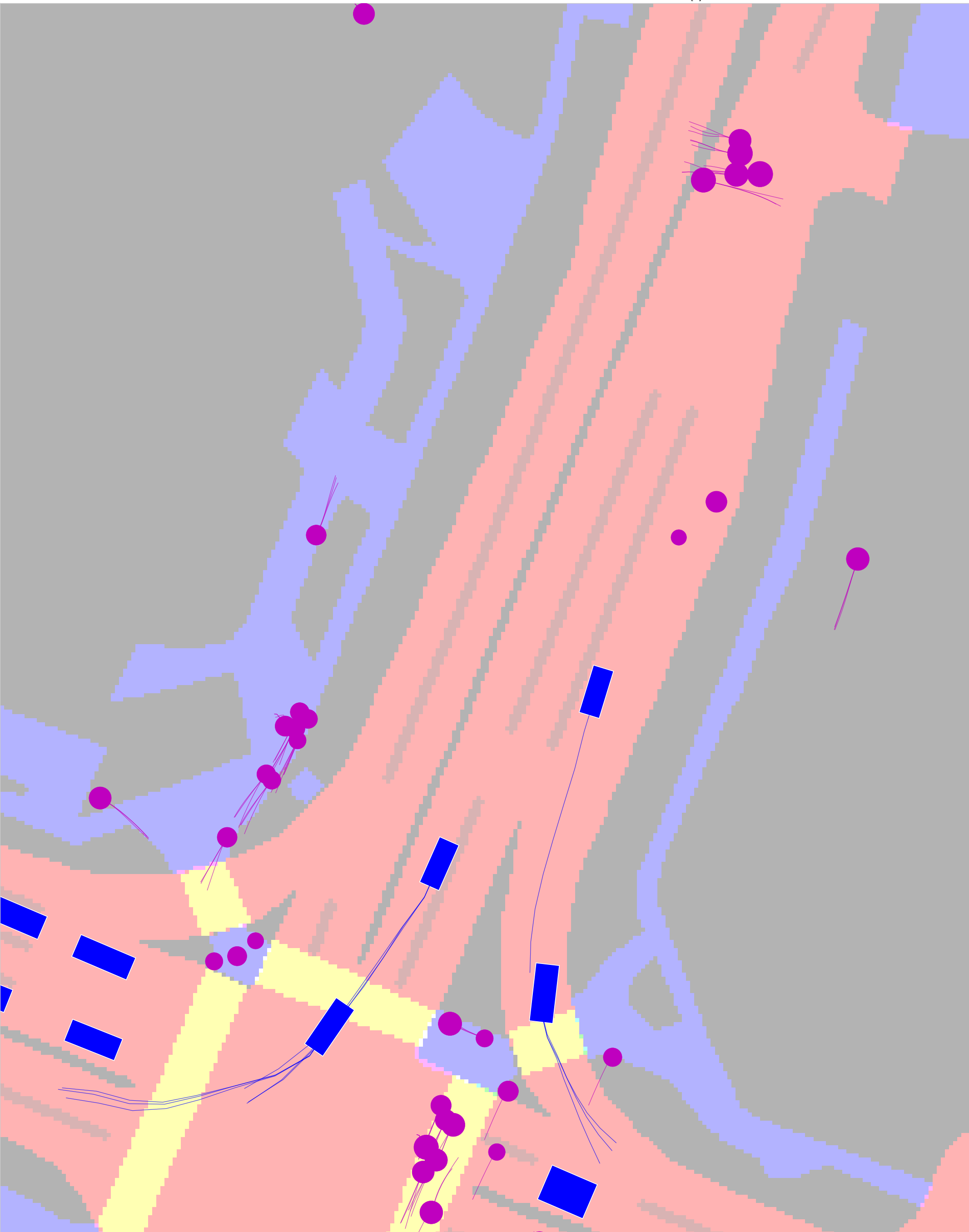}
    \caption{Visual comparison of ground truth (left) and prediction trajectories (right) for a particular scene for both vehicles (blue rectangle) and pedestrians (pink circle). We can observe that each agent has multimodal predictions, with high-confidence predictions aligning closely with the ground truth paths}
    \label{fig3}
\vspace{-0.25in}\end{figure}

As mentioned earlier, the focal mode of visual cognition involves understanding the meaning and significance of the perceived elements in the environment as the first step for determining the next action. Specifically, humans imbibe critical cues from the perceived elements based on their knowledge, past experiences, and language understanding. For example, according to human cognition, it is trivial to conclude that two agents can not actively interact with each other if there are physical barriers present between them. But this intuitive fact is based on semantic information about the physical barriers. Hence, we implement this technique to detect barriers between agents to avoid clique formation between completely irrelevant agents. 
To understand the formation of cliques better, let us consider the scene in Figure ~\ref{fig:clique}(3).
In this scene, agents A and B are moving toward each other, but there is no possible interaction between the agents due to the physical barrier (road divider) between them. Hence these agents need not be considered together in the same clique.  To check if there are any barriers between the two agents, map semantics can be used which are derived from scene segmentation. Given these map attributes as line segments dividing roads, we check if there are dividers between any two agents. Consider a line segment $\tau_{A \rightarrow{B}}$ drawn between the agents $A$ and $B$,$\tau_{C \rightarrow D}$ drawn between $C$ and $D$ in Figure ~\ref{fig:clique}(3) respectively. The line segments $\tau_{A \rightarrow{B}}$ and $\tau_{C\rightarrow D}$ intersect if $ACD$ and $BCD$ have opposite directions meaning either $ACD$ or $BCD$ is counterclockwise but not both. To check the direction of $ACD$, we calculate the slopes $m_{AC}$ and $m_{AD}$ of $AC$ and $AD$, respectively.  If $m_{AC} < m_{AD}$, then $ACD$ has a clockwise direction, else otherwise. We use this condition to check 
if two agents will intersect with each other.
\subsubsection{Overlapping lanes}
One of the techniques involved in the focal mode of visual cognition includes estimating the future possible trajectories based on lane semantics. In other words, the interaction between agents can also be determined based on the lanes which are available in the future for each agent. Based on this information, we can track if there will be any possible intersection of lanes between two agents for them to have an interaction. Lanes for each agent can be found using lane tracking algorithms \cite{NGUYEN2018822}\cite{lane}. They can also be derived from map information assuming the data is accurate. Considering the map semantics to be accurate, we track the possible lanes of each agent. We have lane dividers spanning all the lanes on the roads. These dividers are defined by a set of token identifiers $\tau_{e_{0:N}}$ and are connected with each other by common end tokens $\tau_{e_0}$ and $\tau_{e_N}$. The direction of traffic in a particular lane divider can be obtained by extrapolating any of the tokens from $\tau_{e_{0:N}}$. Each agent can be associated with a lane divider $\tau_{d_i}$ based on its current position. To find if any two agents would meet in the future, we find if there's any intersection in the list of possible lanes of each agent $\tau_{l_i}$. In order to find $\tau_{l_i}$ for each agent $i$, we collect all lane dividers connected to $\tau_{d_i}$ by considering the dividers having $\tau_{e_N}$ as a common token. Consequently, we check for the intersection of the list of lane tokens between agents, $\alpha_{ij} = \mathbb{I}(\tau_{l_i} \cap \tau_{l_j})$.

\subsection{Multi-head Attention-based Global Encoding}
Once the cliques are formed, their local semantic maps and node state and edge history encodings are extracted and further processed through a multi-head attention mechanism to obtain global encodings. To understand the significance of this mechanism, let us consider the state-of-the-art architecture for trajectory prediction consisting of a CVAE model that encodes node history and local map information through a Gibbs distribution and passes through a policy network to generate trajectory predictions. Since agents follow a regular pattern and are highly correlated with each other, global map and state history information is crucial to capture the joint trajectory predictions. Specifically, in cases where agents are occluded from the ego vehicle (cross-traffic scenario), local map information and its associated state history information are not sufficient to capture the global information of the agents in the clique. Also, global context is required to capture the inter-clique information essential for minimizing the collision rate between cliques. Hence, we propose a multi-head attention mechanism by passing map encodings, node history, and edge history information as unique heads through a custom-designed transformer network. The model captures different dependencies and relationships between these heads, thereby enabling active interaction between ambient and focal modes of the vision system. Let the map $M^t_i$ of each agent $i$ be encoded via CNN to produce local encodings $M_{l_i}^t$, and similarly, state and edge histories $(h_{s_i}^t, h_{e_i}^t)$ be encoded into $(h_{s_{l_i}}^t, h_{e_{l_i}}^t )$ using LSTMs. The multi-head attention mechanism acts on these local features ($M_{l_i}^t,h_{s_{l_i}}^t, h_{e_{l_i}}^t $) to produce global features  ($M_{g_i}^t, h_{s_{g_i}}^t, h_{e_{g_i}}^t $) and appended with the local contextual information to form ($\vec M^t_i, \vec h_{s_i}^t, \vec h_{e_i}^t $).  
 The multi-head attention mechanism structure consists of a fully connected layer $FC$ outputting the weights of an attention matrix $A$ populated on its non-diagonal elements.  In greater detail, $FC$ layer outputs a probability weight for all agents with respect to each and every other agent. The local features are then multiplied with the weights to get a global feature vector of the same dimension,
$M_{g_i}^t=\sum w_{i, j} M_{l_i}^t$ , 
$h_{s_{g_i}}^t=\sum w_{i, j} h_{s_{l_i}}^t$ and  $ h_{e_{g_i}}^t = \Sigma w_{i,j}  h_{e_{l_i}}^t$. 

\begin{figure}
\centering
\includegraphics[width=0.49\textwidth]
   {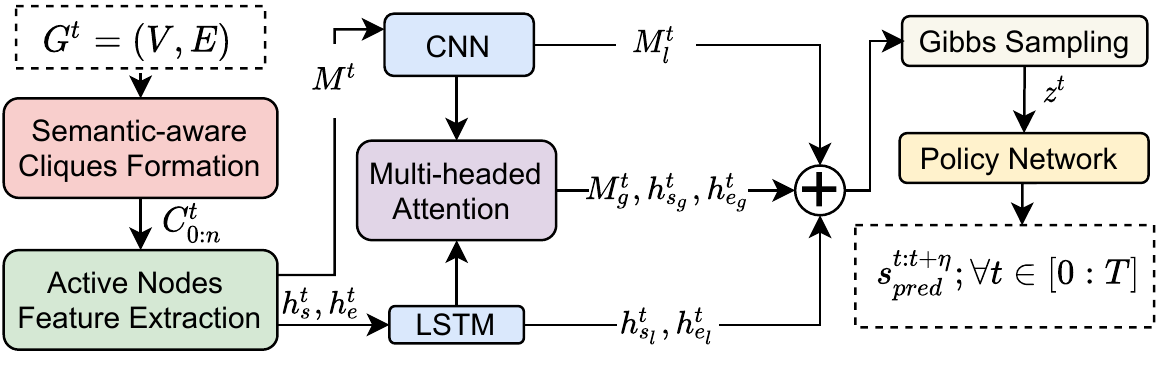}\vspace{-0.1in}
   \caption{
   Graph is generated with nodes as agents. Cliques are formed based on various techniques mentioned in Section 3.1, further passed on to get batch cliques containing node histories and local maps. Upon further processing, they are fed into the Multimodal attention mechanism to get global and local encodings. These encodings are fed into a policy network to obtain closed-loop trajectory predictions.}
   \label{fig:simf}
\vspace{-0.24in}\end{figure}
\subsection{Temporal Latent Encodings}
In SIMMF, we present a method that utilizes both modes of interaction to capture dynamic information by associating each node with a time-dependent latent variable. To achieve this, we encode the 
global information along with local information from both modes of visual perception into discrete latent variables 
$z^t_i = [z^t_1, z^t_2, z^t_3, ..., z^t_N]$ 
for each clique $C^t_{1:n}$ consisting of agents 1 to $N$. To elaborate, consider the temporal encoding framework as a function $g$. We can write 
$z_i^{t} = g(\vec h_{s_i}^{t},\vec h_{e_i}^{t}, \vec M_i^t,s_{{ref}_i}^{t:t+\eta})$ where $s_{ref}$ denotes the reference trajectories during training, and the former variables represent the local and global encodings concatenated together. Note that this is contrary to recent work that capture temporal information implicitly by incorporating influence from neighboring nodes using edge encoders \cite{salzmann2021trajectron}, which incorporates only the ambient mode of information. 
Consequently, the temporal information is ignored if the latent encodings are not iteratively captured based on timestamps, hurting prediction accuracy as indicated by our experimental results. The addition of the time-dependent latent variables has significantly improved the optimality in choice of agents. This can be seen in dense scenes, where the larger scene graphs are partitioned into small cliques. \vspace{-0.05in}\subsection{Overall pipeline}
In SIMMF, we generate a spatio-temporal graph $G^t$, with nodes representing agent and edges representing their interactions evaluated using the techniques mentioned in Section 3.1. The degree of interaction $\alpha$ is populated in an adjacency matrix to form optimal cliques $C^t$, i.e., sub-graphs of agents which have considerable interaction with each other. These cliques are further processed to obtain the active nodes and their relative states in each clique. Consequently, they are rearranged into a list of active nodes containing their corresponding relative state and edge histories, i.e., the batches are populated with the aggregated information of all the nodes along with their corresponding clique details. Further, we generate local semantic maps \cite{caesar2020nuscenes} $M^t$, by rotating and cropping according to each agent's heading and type. These local maps are passed on to CNN to get local map encodings $M_l^t$,  while the state and edge histories are encoded into the feature vectors $(h_{s_l}^t, h_{e_l}^t)$ via LSTMs. These vectors are passed through the Multi-head attention-based mechanism to obtain global encodings  ($M_g^t, h_{s_g}^t, h_{e_g}^t $). The global encodings appended with the local encodings to form ($\vec M^t, \vec h_{s}^t, \vec h_{e}^t $) containing the complete information of the scene, further passed through the Gibbs discrete sampling to produce latent variable encodings for each timestamp $0:T$ to capture the temporal information. These time-dependent latent variables $z^t$ are passed through the decoder containing the policy network to obtain the joint trajectory predictions $s_{pred}^{t:t+\eta}$, as illustrated in Figure \ref{fig:simf}.
\section{Experimental Results}
\subsection{Evaluation Metrics and Baselines} 
We use the standard evaluation metrics, also employed by the prior methods \cite{chen2022scept, salzmann2021trajectron, agentformer, laformer} namely Average Displacement Error (ADE) and Final Displacement Error (FDE). These metrics are based on $L2$ distance. The former is between the ground truth and predicted trajectories, whereas the latter is between the predicted and ground truth final agents' positions associated with the given time stamps. In addition, we also introduce a new metric, named Mean Average Count (mAC) which is inspired by a metric in \cite{waymo}. The mAC is determined as follows: We compute the trajectory prediction of the highest confidence by choosing the Best-of-N extension from the encoder output, i.e. we take the best from the N highest probability modes of the latent variable encodings taken at a time. Next, we check if the prediction is within lateral and longitudinal thresholds $\eta/3\mu(v_x)$ and $\eta/3\mu(v_y)$, respectively, where $\eta$ is time length and $\mu$ is a scaling factor computed based on the velocities $v_x$ and $v_y$. Finally, if the prediction lies within the thresholds, then it is considered a true positive; otherwise considered a false positive. The average of false positives is calculated per clique to estimate the accuracy of each clique. The overall mAC is calculated as the mean of all the cliques in the scene. For all metrics, the lower is better, and bold/italic marks the best/second-best value.
\begin{table}[h]
\begin{center}
\begin{tabular}{ccccc}
\toprule
Method  &@1s & @2s & @3s & @4s  \\
\midrule
S-LSTM [24]& 0.47 & - & 1.61 & - \\
CSP[25]      & 0.46 & - & 1.50 & - \\
CAR-Net [26] & 0.38 & - & 1.35 & - \\
SpAGNN [27] & 0.35 & -& 1.23 & - \\
Trajectron++ [19] & \textbf{0.07} & \textbf{0.45} & \textit{1.14 }& 2.20 \\
ScePT(Best-of-3) & 0.40 & 0.80 & 1.36 &\textit{ 2.14} \\
\midrule
SIMMF(Best-of-3) & \textit{0.32 }& \textit{0.48} & \textbf{0.72}
 & \textbf{1.18 }\\
\bottomrule
\end{tabular}
\end{center}\vspace{-0,1in}
\caption{Comparison of FDE (in meters) with Baselines. Ours perform better among most of the baselines, specifically in the latter timesteps of the prediction horizon.}\label{fde}  \vspace{-.2in}
\end{table}
\begin{table}[h]
\begin{center}
    \begin{tabular}{ccccc}
   
\toprule
\multirow{ 2}{*}{\# samples}  &
\multicolumn{2}{c}{ADE} & \multicolumn{2}{c}{mAC}\\
\cmidrule{2-5}
 & ScePT  & Ours & ScePT  & Ours\\
\midrule

 2 &
0.35/0.11 & \textbf{0.22/0.07} & 0.33  & \textbf{0} \\

 3 &
0.31/0.11 & \textbf{0.26/0.07} & 1.33  & \textbf{0}

\\
 5 & 0.31/0.07 & \textbf{0.17/0.06} &
\textbf{0.33}  & 0.67
\\ 10 & 0.35/0.07 & \textbf{0.27/0.03}
& 2 &\textbf{ 1.67}
\\

\bottomrule
\end{tabular}

\end{center}
\vspace{-0.1in}
\caption{Comparison of ADE (in meters) and mAC (in percentage) with ScePT. SIMMF performs better for all samples for ADE, for most samples for mAC.}\label{ade}\vspace{-0.1in}
\end{table}
\begin{table*}[h]
\centering
  \begin{tabular}{ccccccccc}
   
\toprule
time stamp  &SIMMF & w/o S & w/o T &  w/o A & w/o B & w/o O & w/o D & ScePT\\
\midrule
@0.5s & \textbf{0.19} & 0.28 & 0.22 & 0.33 & 0.19 & 0.20 & 0.31 & 0.33 \\
 @1.5s  &\textbf{ 0.41} & 0.50 & 0.46 & 0.55 & 0.43 & 0.44 & 0.50 & 0.54 \\

  @2.5s&\textbf{0.56}& 0.69&0.62& 0.76 & 0.60 & 0.61 & 0.68 & 0.76\\
@3.5s & \textbf{0.93} & 1.08 & 0.99 & 1.21 & 0.92 & 0.92 & 0.76 & 1.22\\
\bottomrule
\end{tabular}
\vspace{0.1in}
\caption{Ablation study of techniques presented in SIMMF according to FDE. Legend: S- Semantic Cliques, T - Temporal encodings, A - Attn mechanism, B - Barrier Detection, O - Overlapping lanes, D - Interaction using direction. $A$ has the highest influence on SIMMF, followed by the parameter $D$ in semantic-aware clique formation.}
\label{table:ablation}
\vspace{-0.1in}\end{table*}

Our results are compared against these baselines: Social LSTM \cite{7780479} where each agent is modeled with LSTMs and followed by the multiagent “Social” pooling layer; Convolutional Social Pooling (CSP) \cite{Deo_2018} which explicitly considers a fixed number of trajectory classes thereby converting prediction to a classification task; CAR-Net \cite{sadeghian2018carnet} solves the prediction task by localizing the regions of interest for each agent using LSTMs-based visual attention; SpAGNN \cite{casas2019spatiallyaware} where raw LIDAR data and semantic maps are encoded using CNN followed by GNN to produce probabilistic trajectories; Trajectron++ \cite{salzmann2021trajectron} and ScePT \cite{chen2022scept} where the agents are represented as a spatiotemporal graph and the predictions are forecasted through CVAE model.
\subsection{Results}
\begin{figure}{
 \centering
 \vspace{0pt}}\includegraphics[width=0.49\textwidth]{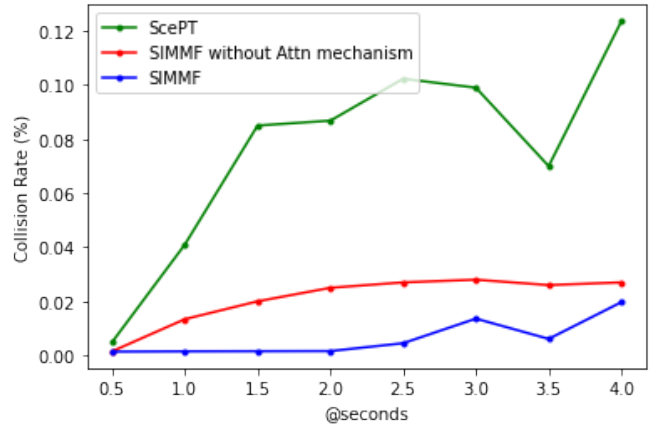}
 \vspace{-0.15in}
\caption{SIMMF demonstrates a reduced collision rate compared to ScePT, and the incorporation of an attention mechanism further decreases the collision rate, resulting in improved scene consistency.}
\vspace{0pt}
\label{fig:collision}
 \vspace{-0.2in}\end{figure}

In this section, we present the comparison and ablation analysis on nuScenes dataset \cite{caesar2020nuscenes}. The ADE and FDE are represented in meters, and mAC is represented as the percentage of misses,  i.e., the percentage of the average number of nodes having predictions that do not lie within the expected threshold region. Furthermore, we evaluate with Best-of-N metric similar to the baselines to establish a fair comparison.
\\
\vspace{0pt}
\textbf{Comparison with baselines:}
In this section, we compare our approach, SIMMF, against the given baselines. Table~\ref{fde} summarizes the FDE-based comparison of all methods. It can be seen that SIMMF gives overall better results among almost all the baselines, especially in the latter timesteps of the prediction horizon. This clearly shows that the temporal dependencies in lateral encodings are vital in reducing the accumulated error from the previous timesteps' predictions. Although Trajectron++ \cite{salzmann2021trajectron} gives the best result in the short term, it is computationally expensive to execute as it does not form cliques. Among baselines, only the ScePT forms cliques; therefore, we center our remaining comparison analysis against the ScePT method. 
Table\ref{ade} compares our method with ScePT using mAC and ADE. The results of ADE (represented for vehicles/pedestrians respectively) show that our model performs better than ScePT. Furthermore, as we increase the number of samples taken from the Gibbs distribution, there is an increase in miss rate, possibly due to the impact of the tunable risk measure incorporated from ScePT that modifies the confidence level of predictions.  
Specifically, there are high-confidence predictions that are different from the ground truth when increasing the number of sample encodings taken at a time. In terms of collision rate, SIMMF shows significantly enhanced scene consistency than ScePT, as seen from the graphs in Figure ~\ref{fig:collision}, demonstrating a considerable decrease in collision rate, specifically with the use of the attention mechanism. Apart from the quantitative study, we qualitatively compare trajectory predictions in comparison with their corresponding ground truth visualizations in Figure ~\ref{fig3}. 
It is evident that the predictions closely align with the ground truth for both vehicles and pedestrians. 
 
\textbf{Ablation study:} 
To develop an understanding of the individual components of SIMMF, we conduct an ablation study having two parts. First, we ablate the model architecture on semantic aware clique formation, temporal encodings, and attention mechanism. Secondly, we investigate the impact of semantic cliques by ablating the various techniques of Section 3.1, including direction, barrier detection, and overlapping lanes. We utilize ScePT for the ablation study since it quantifies the degree of interaction solely using the Euclidean distance parameter. Table~\ref{table:ablation} summarizes our results, columns 2-5 and 5-9 highlight the first and second parts of our ablation study. It can be observed that the attention mechanism has the greatest impact on the accuracy, whereas the temporal factor has more impact on the later timesteps. From the second part, we can ascertain that the direction factor has the greatest impact, owing to its intuitive reasoning of relative velocities. 
\section{Limitations and Future Work}
One of the limiting factors of our approach is that we assume road semantics information is available and accurate, which might not be the case in real-world settings. Therefore, our future work includes working on an end-to-end model using only camera inputs to provide solutions to vehicles that lack expensive onboard sensors for feature extraction. Secondly, our method implicitly captures inter-clique interaction, which with explicit modeling, may lead to further improvements in our performance. Therefore, another of our future works is explicitly capturing inter-clique interactions that may be achieved using reparameterization techniques to backpropagate through the time-dependent latent variables sampled during training for temporal latent encodings. 
\bibliography{example}

\end{document}